\documentclass[10pt,twocolumn]{article}

\usepackage{cvpr}
\usepackage{times}
\usepackage{epsfig}
\usepackage{graphicx}
\usepackage{amsmath}
\usepackage{amssymb}
\usepackage{mathrsfs}
\usepackage{multirow}
\usepackage{color}
\usepackage{subcaption}

\usepackage[pagebackref=true,breaklinks=true,letterpaper=true,colorlinks,bookmarks=false]{hyperref}

 \cvprfinalcopy 

\begin{document}

\title{Adaptive Context-Aware Multi-Modal Network for Depth Completion}

\author{\mbox{Shanshan Zhao\quad\quad Mingming Gong \quad\quad Huan Fu \quad\quad Dacheng Tao}\\
}

\maketitle

\begin{abstract}
{
Depth completion aims to recover a dense depth map from the sparse depth data and the corresponding single RGB image.
The observed pixels provide the significant guidance for the recovery of the unobserved pixels' depth. However, due to the sparsity of the depth data, the standard convolution operation, exploited by most of existing methods, is not effective to model the observed contexts with depth values.  
To address this issue, we propose to adopt the graph propagation to capture the observed spatial contexts. Specifically, we first construct multiple graphs at different scales from observed pixels. Since the graph structure varies from sample to sample, we then apply the attention mechanism on the propagation, which encourages the network to model the contextual information adaptively. Furthermore, considering the mutli-modality of input data, we exploit the graph propagation on the two modalities respectively to extract multi-modal representations.
Finally, we introduce the symmetric gated fusion strategy to exploit the extracted multi-modal features effectively.
The proposed strategy preserves the original information for one modality and also absorbs complementary information from the other through learning the adaptive gating weights. Our model, named Adaptive Context-Aware Multi-Modal Network (ACMNet), achieves the state-of-the-art performance on two benchmarks, {\it i.e.}, KITTI and NYU-v2, and at the same time has fewer parameters than latest models. Our code is available at: \url{https://github.com/sshan-zhao/ACMNet}.
}
\end{abstract}

\section{Introduction}
Depth information is crucial for 3D vision tasks, {\it e.g.}, 6D object pose estimation~\cite{wang2019densefusion}, 3D object detection~\cite{Xu_2018_CVPR1}, and human pose estimation~\cite{Moon_2018_CVPR}. To complete these tasks, various depth sensors such as LiDAR have been invented to acquire depth information. However, current depth sensors are not able to obtain dense maps for outdoor scenes, which are essential in various applications, especially autonomous driving. Therefore, depth completion from sparse depth maps\footnote{The sparse depth map is generated by projecting the LiDAR data to the image plane, and the value in locations without depth information is $0$.} and RGB images has attracted intensive attention.
\begin{figure*}
\begin{center}
 \includegraphics[width=10cm,height=1.65in]{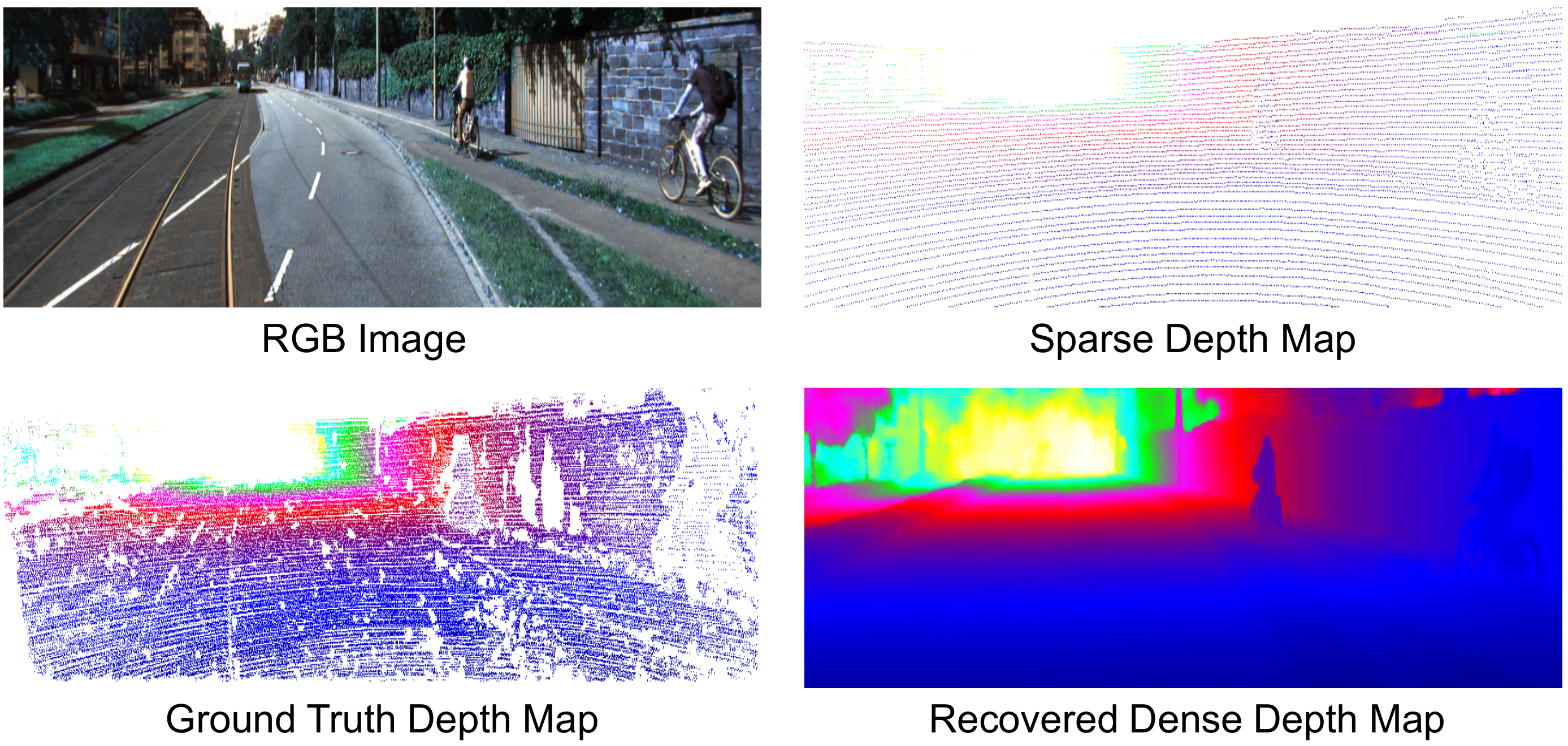}
\end{center}
   \caption{Depth Completion from LiDAR Data and RGB Image by ACMNet. Top: RGB image and sparse LiDAR data; Bottom: ground truth depth map and dense depth map obtained by our approach.}
\label{fig:firstpage}
\end{figure*}
Depth completion is a challenging problem because the depth values obtained by sensors are highly sparse and irregularly spaced. For example, in the KITTI dataset~\cite{Geiger2012CVPR}, there are only $5.9\%$ pixels with depth information obtained by the Velodyne HDL-64e (64 layers) LiDAR in the whole image space, as shown in Figure~\ref{fig:firstpage}. Traditional methods~\cite{ferstl2013image,herrera2013depth,schneider2016semantically} rely on handcrafted features and global constraints on the output depth values, which are inaccurate.
Recent studies ~\cite{Zhang_2018_CVPR,uhrig2017sparsity,mal2018sparse,jaritz2018sparse,Imran_2019_CVPR,Atapour-Abarghouei_2019_CVPR,Cheng_2019_CVPR,Chen_2019_ICCV,zhong2019deep,Eldesokey_2020_CVPR,Lu_2020_CVPR}
have demonstrated great advantages of deep Convolutional Neural Networks (CNNs) on depth completion. By extending the convolutional operation with sparsity-invariance~\cite{uhrig2017sparsity,8946876,eldesokey2018confidence} or introducing more geometric information~\cite{qiu2018deeplidar,Xu_2019_ICCV}, these deep methods can achieve way better performance than traditional methods.

\begin{figure}[t]
\begin{center}
\includegraphics[width=0.8\linewidth,height=1.4in]{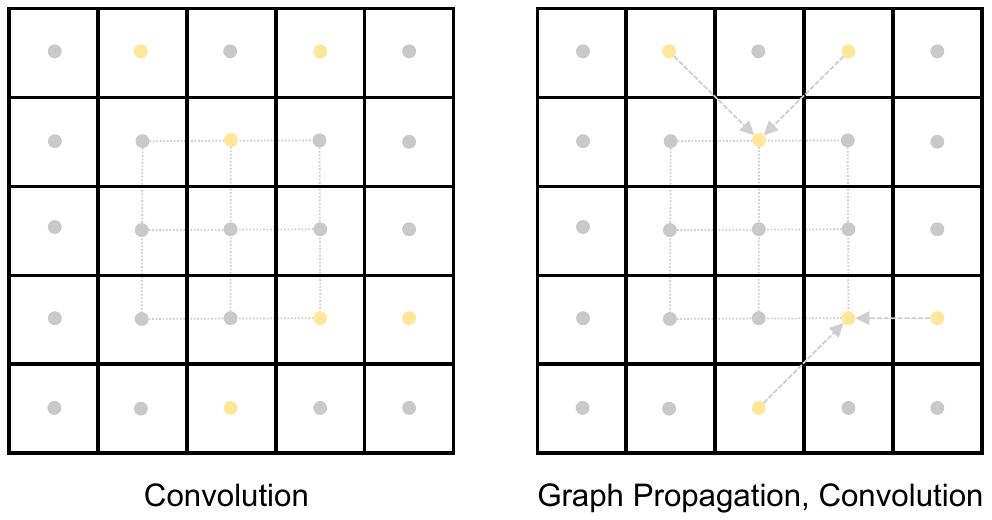}
\end{center}
   \caption{Illustration of convolution and graph propagation. Left: convolution ($3\times 3$ kernel); Right: graph propagation (2-nearest neighbours) and convolution ($3\times 3$ kernel). The observed pixels are marked by the yellow, while the unobserved are marked by the gray.}
\label{fig:conv_gp}
\end{figure}
In spite of the encouraging progress, existing depth completion methods suffer from a significant issue, which limits the depth completion performance. Specifically, the conventional convolutional operation applies kernels with regular structure ({\it e.g.,} $3\times3$) at all locations, which ignores the fact that the observed depth values are irregularly distributed in a sparse depth map and associates limited observed contexts for the unobserved, as shown in Figure~\ref{fig:conv_gp}.
Thus, CNN-based methods are not adaptive to the pattern of observed spatial contextual information in a sparse depth map, resulting in a sub-optimal prediction of depth in unobserved locations. 

To address this issue and further boost depth completion accuracy, we propose an Adaptive Context-Aware Multi-Modal Network (ACMNet, shown in Figure~\ref{fig:framework}). 
Firstly, inspired by recent works on point cloud analysis~\cite{wang2018dynamic}, we model the observed contextual information adaptively by applying attention based graph propagation within multiple graphs constructed from observed pixels. Based on the efficient graph propagation, the model can associate the spatial context with observed depth values and then enhance the features of the unobserved pixels. To illustrate this, we provide a simple example in Figure~\ref{fig:conv_gp}. Compared to the sole convolutional operations, the proposed graph propagation (followed by a convolution) can make the unobserved pixels capture more related observed contextual information.

Furthermore, since we have multi-modality data, we need to reconsider the novel graph propagation in a multi-modal setting. 
Firstly, to better learn the relationship between observed pixels (nodes), we use the co-attention mechanism~\cite{lu2016hierarchical} to propagate the multi-modal information of observed pixels in a symmetric structure. This step is conducted in the encoder to extract multi-scale and multi-modal features. However, this mechanism does not consider the fusion of multi-modal contextual information. A simple way to fuse the multi-modal data is by applying the simple concatenation or element-wise summation operation on the extracted feature maps, which was used by most of the existing works, {\it e.g.,} \cite{jaritz2018sparse, qiu2018deeplidar}. However, this type of fusion strategy cannot fully explore the heterogeneity of the two modalities.
To address the issue, we further present the symmetric gated fusion strategy to combine the depth and RGB information in the decoder. In specific, the presented fusion strategy consists of two branches. One branch focuses on fusing the RGB information as supplementary into the depth information through learning an adaptive gating function, and the other one does the opposite. Therefore, each branch can maintain its own information and benefit from supplementary information from the other.
Benefiting from the adaptive co-attention guided graph propagation and symmetric gated multi-modal feature fusion, our ACMNet is able to generate high-quality dense depth maps.
In summary, our main contributions are:
\begin{itemize}
\item We introduce the co-attention guided graph propagation to our depth completion network, which is adaptive to the sparsity patterns of sparse depth input and thus enables the unobserved pixels to capture useful observed contextual information more effectively.
\item To fuse the multi-modal contextual information efficiently, we further present the symmetric gated fusion strategy, which can learn the heterogeneity of the two modalities adaptively.
\item We demonstrate the effectiveness of ACMNet on two benchmarks, {\it i.e.,} KITTI Depth Completion Dataset~\cite{Geiger2012CVPR} and NYU-v2 Dataset~\cite{silberman2012indoor}.
\end{itemize}

\section{Related Work}
{\bf Depth Completion.} Traditional approaches solve the depth completion problem by formulating the task as an energy function optimization problem~\cite{ferstl2013image,barron2016fast,herrera2013depth,schneider2016semantically}. However, these works showed some limitations in
performance due to the employment of
hand-crafted features.

Currently, CNNs have been a dominant solution for depth completion~\cite{qiu2018deeplidar,chodosh2018deep,cheng2018depth,tang2019learning,van2019sparse,ma2018self,yang2018conditional,8946876,atapour2019complete,cheng2020cspn++,Eldesokey_2020_CVPR,Lu_2020_CVPR,liao2017parse,li2020multi}, outperforming traditional methods by a wide margin. In specific, to learn representations of the irregular and sparse LiDAR data, Uhrig {\it et al}.~\cite{uhrig2017sparsity} proposed the sparsity-invariant convolutional operation. Following this work, some variants of the sparse convolution are introduced~\cite{eldesokey2018confidence,8946876,Eldesokey_2020_CVPR}.
In the case of additional RGB data,
Jaritz {\it et al}.~\cite{jaritz2018sparse} showed that the late fusion strategy outperformed the early fusion. Ma {\it et al}.~\cite{ma2018self} utilized self-supervised learning on sparse LiDAR data coupled with the stereo image pair to mitigate the need for ground truth dense depth. Yang {\it et al.}~\cite{Yang_2019_CVPR} exploited the Conditional Prior Network~\cite{yang2018conditional} to learn a depth prior on synthetic images.
Additionally, there are also a bunch of works~\cite{Zhang_2018_CVPR,van2019sparse,Xu_2019_ICCV} exploring other cues. For example,
Zhang {\it et al}.~\cite{Zhang_2018_CVPR} trained a network to predict local surface normals for indoor scene depth completion, and later an extension for outdoor scenes was introduced in their latest work~\cite{qiu2018deeplidar}. Similarly, Xu {\it et al}.~\cite{Xu_2019_ICCV} also explored the surface normal information to improve the performance by introducing a diffusion module. Cheng {\it et al.}~\cite{cheng2018depth,cheng2020cspn++} proposed to learn affinities between adjacent pixels for the spatial propagation of the depth information. Following the two works, a recent work~\cite{park2020non} improved the propagation strategy through concentrating on the non-local neighbors and introducing a learnable affinity normalization. 
Inspired by the guided image filtering, Tang {\it et al.}~\cite{tang2019learning} designed a guided convolution module, which generates dynamic spatially-variant kernels using the image features, to extract the depth image features. In comparison, a recent work~\cite{xiong2020sparse} proposed to dynamically learn the filter by applying the Graph Neural Network (GNN)~\cite{zhou2018graph} on the graph constructed from the predicted dense depth map.
In contrast to these approaches, which paid little attention to the modelling of the multi-modal contexts, our work mainly aims at making
unobserved pixels capture more useful observed contextual information from the input multi-modal data. Additionally, it is worth pointing out that although the latest work~\cite{xiong2020sparse} also exploits the graph models, there are many differences between it and ours. For example, it aims to 
consider the neighborhood relationship of the points in the 3D space through constructing a 3D graph from the dense depth map, which is obtained using a deep model. To arrive at this, it applies the dynamic kernel, which is learned through using a typical GNN model on the constructed graph, on the dense features at $1/8$ of original scale. In comparison, 
in this paper we study the propagation of the contexts with observed depth values at multiple scales in a multi-modal setting to enhance the features of the unobserved pixels. 

{\bf Monocular Depth Estimation.} From approaches based on probabilistic graphical models ({\it e.g.}, MRFs) with hand-crafted features~\cite{saxena2006learning,saxena2009make3d} to the deep learning-based~\cite{fu2018deep,liu2016learning,godard2017unsupervised,eigen2014depth,zhou2017unsupervised,garg2016unsupervised,8360460,8359371,8506384,8972902,9099063,8331148}, the improvement of performance for monocular depth estimation has been pushed forward. Eigen {\it et al.}~\cite{eigen2014depth} were the first to develop deep models for depth estimation.  Following their work, a lot of supervised approaches~\cite{laina2016deeper,liu2016learning1,fu2018deep,eigen2015predicting} have been proposed. However, these methods rely on large quantities of ground truth depth data, which is hard to acquire. To address this issue, Garg {\it et al.}~\cite{garg2016unsupervised} and Godard {\it et al.}~\cite{godard2017unsupervised} proposed to predict depth maps from stereo pair images by exploring unsupervised cues, while some recent works tried to utilize synthetic data~\cite{atapour2018real,zhao2019geometry,zheng2018t2net,nath2018adadepth,PNVR_2020_CVPR} based on the domain adaptation technique~\cite{pan2009survey}.

{\bf Graph-based Models.} Conventional deep learning modules, such as CNNs, do not perform well on graphs. To model the graph data efficiently, Graph Models have been applied on various computer vision tasks, such as action recognition~\cite{shi2019skeleton,yan2018spatial}, point cloud analysis~\cite{wang2019graph,wang2018dynamic}, few-shot image classification~\cite{garcia2017few,kim2019edge}, and person re-identification~\cite{wu2020adaptive}. Graph Models are able to learn the representation of each target node by propagating its neighborhood information in a data-driven way and thus associate the contextual information. In this work, we design an attention-based graph propagation module and then extend it to the co-attention guided graph propagation for multi-modal data, which is capable of learning an efficient multi-modal representation for the input data through encouraging the adaptive contextual interactions.

{\bf Multi-modal Information Fusion.} Multi-modal information fusion has been studied in various computer vision tasks, such as visual question answering~\cite{ben2017mutan}, video action recognition~\cite{simonyan2014two}, 3D object detection~\cite{yoo20203d}, and many more. A simple approach to fuse the multi-modal data is applying concatenation or summation operation into the input data or extracted feature maps~\cite{jaritz2018sparse,simonyan2014two}. However, for a specific task, different modalities often provide different information, and therefore, the naive fusion strategy might fail to combine them effectively. To address this issue, some works, {\it e.g.,}~\cite{yoo20203d,hori2018multimodal}, proposed to exploit the attention mechanism to improve the performance. As for depth completion, current works mainly employed the naive fusion strategy. In fact, both naive strategy and attention based approaches  fuse  the  multi-modal  features  in  a  single  way, which is not enough to extract complementary information and then limits the performance. 
In contrast, we present the symmetric gated fusion strategy consisting of two fusion paths, each of which only focuses on one modal and extracts useful information adaptively from the other.

\section{Our Approach}
\subsection{Problem Formulation}
\begin{figure*}
\begin{center}
 \includegraphics[width=0.97\linewidth,height=3.5in]{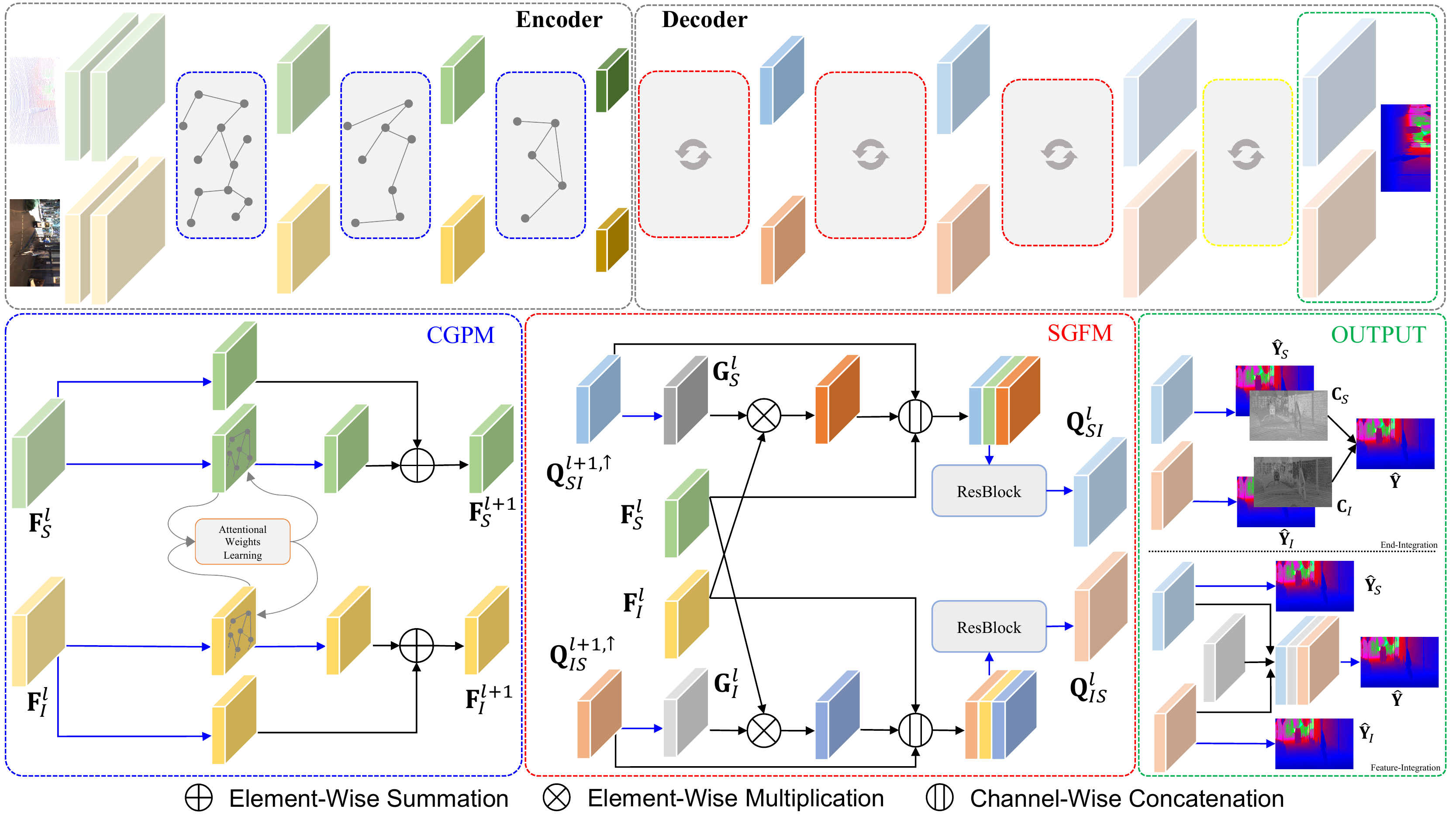}
\end{center}
   \caption{The proposed ACMNet in this paper. Left upper part: Encoder; Right upper part: Decoder. In encoding stage, we extract multi-scale multi-modal features using a stack of {\color{blue}CGPMs} (Marked by blue dotted box, Sec.~\ref{sec:gp}), and the adaptive attentional weights are learnt from spatial locations, depth features and RGB features. In decoding stage, we fuse the multi-modal features progressively by exploiting the {\color{red}SGFMs}, represented by red dotted boxes (Sec.~\ref{sec:cgf}). Lastly, final {\color{green}output} is calculated from the dense maps and confidence maps produced by the two branches of the decoder or predicted using the intermediate fused features maps, shown in the green dotted box (Sec.~\ref{sec:branch_fusion}). Note that, the yellow dotted box denotes that there is no ResBlock behind the initial fusion (see Sec.~\ref{sec:cgf}) in the SGFM. Blue arrow: convolution; Gray arrow: graph propagation; Black arrow: summation/multiplication/concatenation.}
\label{fig:framework}
\end{figure*}

Our goal is to recover a dense depth map from the observed sparse depth data and a single RGB image. Mathematically, given a set of paired samples $\{({\bf X}_S, {\bf X}_I)_i\}^{N-1}_{i=0}$, we expect to learn a mapping function $f(\cdot)$ that satisfies ${\bf Y} = f({\bf X}_S, {\bf X}_I)$, where ${\bf X}_S\in \mathbb{R}^{H \times W}$, ${\bf X}_I\in \mathbb{R}^{3\times H \times W}$, and ${\bf Y}\in \mathbb{R}^{H\times W}$ represent the sparse depth map, the RGB image, and the ground truth depth map, respectively. To achieve this target, we develop a high-performing depth completion network (ACMNet) building on two novel modules, including a co-attention guided graph propagation module (CGPM) and a symmetric gated fusion module (SGFM), as shown in Figure~\ref{fig:framework}. In specific, we first employ a series of CGPMs to effectively extract contextual information from ${\bf X}_S$ and ${\bf X}_I$.  Then we exploit SGFMs to learn the complementarity between contextual representations from multi-modalities. In the following, we will present our network architecture and the proposed modules in detail.
\subsection{Network Architecture}
\label{sec:network}
Our overall network architecture follows a two-stream encoder-decoder fashion as previously
~\cite{van2019sparse,ma2018self,jaritz2018sparse,Atapour-Abarghouei_2019_CVPR}, but with the improvement by integrating the novel CGPM and SGFM. We show the whole framework in Figure~\ref{fig:framework}, and briefly explain the encoder and the decoder right here.

{\bf Encoder}. The encoder targets learning discriminative multi-scale features from both the sparse depth and the RGB image. While researchers reached a consensus that standard convolutional operations can perform well in the image data, how to extract rich information from observed spatial contexts is still an open problem due to the extreme sparsity~\cite{uhrig2017sparsity,eldesokey2018confidence,8946876,jaritz2018sparse,van2019sparse,tang2019learning,Xu_2019_ICCV}.
In this paper, we show that the proposed CGPM has the potential to capture the related contextual information from the observed pixels with various patterns in an adaptive manner through learning dynamic weights of the relationship between adjacent nodes in the constructed graph. Specifically, our encoder consists of two conventional convolutional layers followed by a stack of CGPMs. The encoded features at each scale $\{{\bf F}_S^l\}_{l=1}^L$ and $\{{\bf F}_I^l\}_{l=1}^L$ can be computed as:
\begin{equation}
\begin{aligned}
&({\bf F}_S^{l},{\bf F}_I^{l}) = f_{e}^{l}({\bf F}_S^{l-l}, {\bf F}_I^{l-1}), {\bf F}_S^l,{\bf F}_I^l\in \mathbb{R}^{C^l\times \frac{H}{2^{l}}\times\frac{W}{2^{l}}}, \\
\end{aligned}
\label{eq:encoder}
\end{equation}
where $l=1,2,...,L$, and $f_{e}^l$ denotes the CGPM at level $l$, and ${\bf F}_S^0$ and ${\bf F}_I^0$ are the outputs of the beginning convolutional layers.

{\bf Decoder.} The decoder aims to predict depth values of unobserved pixels in ${\bf X}_S$ given multi-scale and multi-modal features generated by the encoder mentioned above. To this end, one of the commonly studied problems is how to take full advantage of multi-modal representations. A straightforward idea is to directly concatenate or sum features progressively at different scales~\cite{jaritz2018sparse,qiu2018deeplidar}. However, as analyzed before, these naive fusion strategies fail to model the complementary information between multiple modalities satisfyingly. To alleviate the issue, we propose an adaptive symmetric gated fusion strategy to fuse the multi-modal contextual representations in a parallel structure. In specific, we design two parallel branches in the decoder, {\it i.e.,} the depth and image branches. The depth branch preserves discriminative information of the sparse depth modality and meanwhile adaptively captures comprehensive information from the image model through learning dynamic gating weights, and vise versa for the image branch. The overall decoder architecture is described as follows. 

As shown in Figure~\ref{fig:framework}, at the beginning of the decoder,  we feed ${\bf F}_S^L$ coupled with ${\bf F}_I^L$ into the first SGFM to generate the fused feature ${\bf Q}_{SI}^{L}$ and ${\bf Q}_{SI}^{L,\uparrow}$, which is acquired by up-sampling ${\bf Q}_{SI}^{L}$ through one deconvolutional layer. At the following levels $l$ from $L-1$ to $0$, ${\bf Q}_{SI}^{l+1,\uparrow}$,  ${\bf F}_S^l$ and ${\bf F}_I^l$ are fed into the SGFM at level $l$ together. Similarly, we can obtain the intermediate features  ${\bf Q}_{IS}^{l}$ in the image branch.
The procedure can be expressed as:
\begin{equation}
\begin{aligned}
&({\bf Q}_{SI}^L, {\bf Q}_{IS}^L, {\bf Q}_{SI}^{L,\uparrow}, {\bf Q}_{IS}^{L,\uparrow})=f^{L}_{d}({\bf F}_S^L, {\bf F}_I^L), \\
&({\bf Q}_{SI}^l, {\bf Q}_{IS}^l, {\bf Q}_{SI}^{l,\uparrow}, {\bf Q}_{IS}^{l,\uparrow}) = f_{d}^{l}({\bf Q}_{SI}^{l+1,\uparrow}, {\bf Q}_{IS}^{l+1,\uparrow}, {\bf F}_S^l, {\bf F}_I^l), \\
\end{aligned}
\label{eq:decoder}
\end{equation}
where $l=L-1,L-2,...,0$, and $f_{d}^l$ represents the SGFM.

Finally, we present two methods, {\it i.e.,} end-integration and feature-integration, to combine the two branches to obtain the final recovered dense depth map, which will be described in detail in Sec.~\ref{sec:branch_fusion}.

\subsection{Co-Attention Guided Graph Propagation (CGPM)}
\label{sec:gp}
\begin{figure}[t]
\begin{center}
\includegraphics[width=0.6\linewidth,height=1.4in]{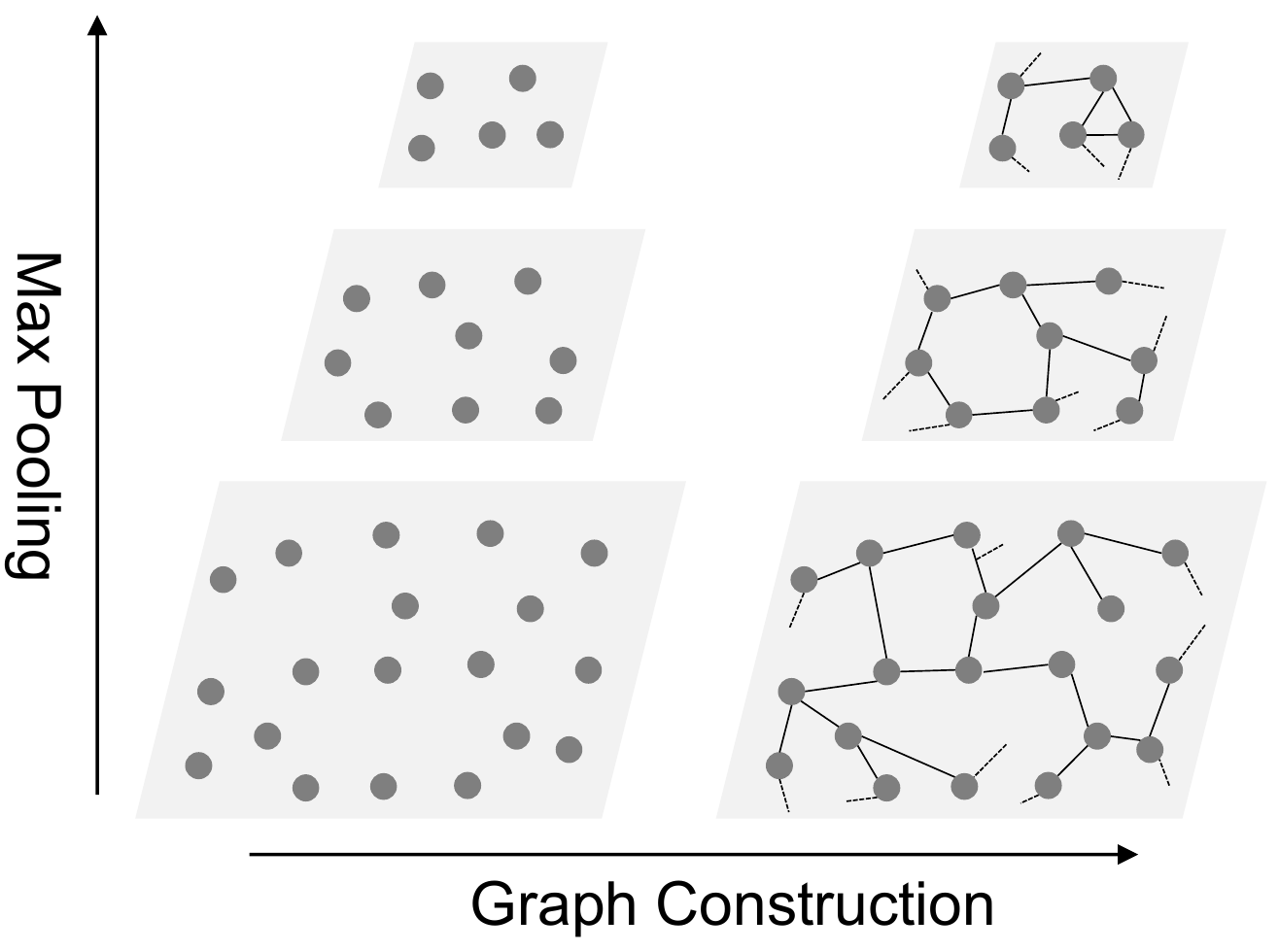}
\end{center}
   \caption{Graph Construction. At each scale, we use $k$ ({\it e.g.,} $k$ is 3 in this example) nearest neighbour to construct the graph from the observed pixels, represented by gray circles.}
\label{fig:graph}
\end{figure}

The proposed CGPM is composed of a residual connection and a co-attention guided graph propagation module.
First, we introduce the basic graph propagation module, which is employed in CGPM.
In specific\footnote{In the following part, we deprecate the scale indexes $l$ to simplify our presentation in some cases.}, given the spatial position set $P=\{p_0,p_1,...,p_{n-1}\}$ of $n$ pixels with observed depth values, we define a graph $G(V,E)$, where $V$ is the vertex (or node) set corresponding to $P$, and $E\subseteq |V|\times|V|$ is the edge set. For a vertex $i$, we connect it to the $k$ nearest neighbour $\mathcal{N}_i$ according to the spatial locations. Note that, we build an individual graph for the CGPM at each scale. Thus, to obtain a specific $P^l$ at level $l$, which is in lower resolution, we generate ${\bf X}_S^l$ by applying max-pooling based down-sampling operation on ${\bf X}_S^{l-1}$. The graph's construction process can be found in Figure~\ref{fig:graph}. In the following, we first introduce the basic attention guided graph propagation component at level $l$ by taking the image stream as an example, then present the full CGPM.

Given the graph $G$ and the input feature maps ${\bf F}_I^{l-1}$, we expect to learn discriminative ${\bf F}_I^{l}$ by both adaptively encoding the contextual information of scenes and exploiting guidance for unobserved pixels from observed pixels. Specifically, we exploit two efficient stages, {\it i.e.,} adaptive feature propagation within observed pixels and feature enhancement of unobserved pixels.

At the first stage, we employ one standard convolutional layer to extract ${\bf F}_I'$ from ${\bf F}_I^{l-1}$, and denote ${\bf F}_{Io}'\in \mathbb{R}^{n\times C}$ as the feature vectors of all the nodes in $G$. Then, we adaptively aggregate neighboured information for each node $i$ in $G$ as:
\begin{equation}
\begin{aligned}
&\alpha^{i,j}=\frac{exp({\bf W}^{i,j})}{\sum_{k\in \mathcal{N}_o^i}exp({\bf W}^{i,k})},\\
&{\bf F}_{Io}^{''i}=\sum_{j\in \mathcal{N}_o^i}\alpha^{i,j}{\bf F}_{Io}^{'i},
\end{aligned}
\label{eq:alpha}
\end{equation}
where $\alpha^{i,j}$ is the computed attentional weight, and ${\bf W}^{i,j}$ is the adaptive weight between neighboured nodes $i$ and $j$. Here, inspired by the works ~\cite{wu2019pointconv,wang2019graph,li2018so} on point cloud, we exploit the self-attention mechanism~\cite{vaswani2017attention} to learn ${\bf W}^{i,j}$ adaptively by modelling the relationship between the connected nodes. Mathematically, the mapping function $f_w$ between ${\bf F}_{Io}'$ and ${\bf W}^{i,j}$ can be expressed as:
\begin{equation}
\begin{aligned}
{\bf W}^{i,j}=f_w([\Delta{p^{i,j}}||\Delta{{\bf F}_{Io}^{'i,j}}]), j\in \mathcal{N}_o^i,
\end{aligned}
\label{eq:dyw}
\end{equation}
where $[\cdot || \cdot]$ represents the concatenation operation, $\Delta{p^{i,j}}=p^j-p^i$ and $\Delta{{\bf F}_{Io}^{'i,j}}={\bf F}_{Io}^{'j}-{\bf F}_{Io}^{'i}$ denote the spatial and feature distances between node $i$ and $j$, respectively. The $f_w$ is implemented by a two-layer MLP, the first one followed by one LeakyReLU activation function~\cite{maas2013rectifier}. Note that, permutation variant operations like convolution are not allowed here due to the unordered input.
After obtaining ${\bf F}_{Io}''$, the features of unobserved pixels are enhanced by a standard convolutional operation. In addition, a residual connection~\cite{he2016deep} is also utilized to preserve early information. We can use the same algorithm to conduct propagation in the depth stream.

As shown in Figure~\ref{fig:framework}, in the CGPM in our encoder, we learn the adaptive weights ${\bf W}_S$ and ${\bf W}_I$ by considering both information from the image stream and the sparse depth stream, inspired by the co-attention mechanism~\cite{lu2016hierarchical}. Therefore, in each CGPM, Eq.~\ref{eq:dyw} can be re-written as:
\begin{equation}
\begin{aligned}
&{\bf W}_S^{i,j}=f_{Sw}([\Delta{p^{i,j}}||[\Delta{{\bf F}_{So}^{'i,j}}||\Delta{{\bf F}_{Io}^{'i,j}}]]), j\in \mathcal{N}_o^i,\\
&{\bf W}_I^{i,j}=f_{Iw}([\Delta{p^{i,j}}||[\Delta{{\bf F}_{Io}^{'i,j}}||\Delta{{\bf F}_{So}^{'i,j}}]]), j\in \mathcal{N}_o^i.
\end{aligned}
\label{eq:dyw2}
\end{equation}

\subsection{Symmetric Gated Fusion (SGFM)}
\label{sec:cgf}
\begin{figure}[t]
\begin{center}
\includegraphics[width=0.8\linewidth,height=3.4in]{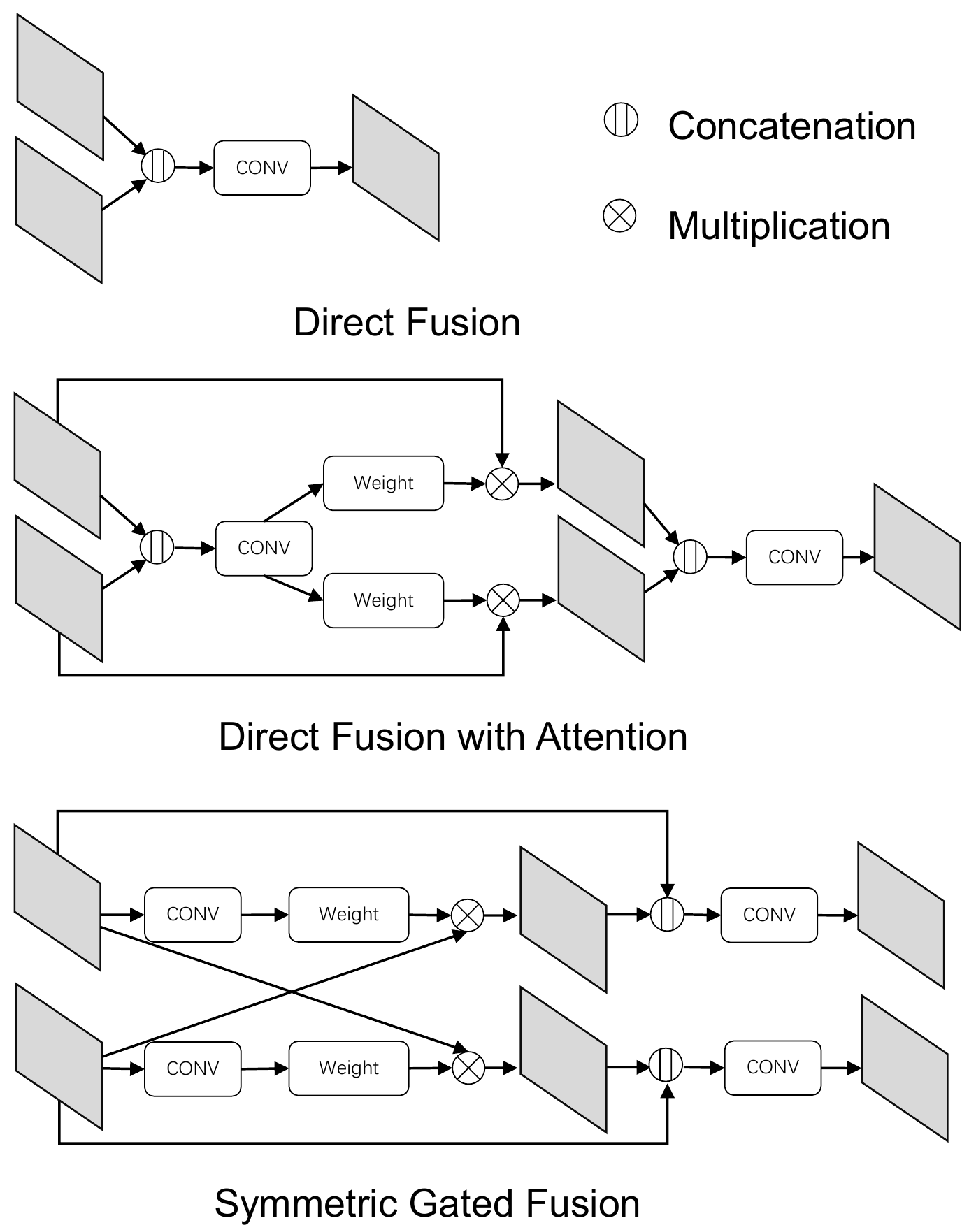}
\end{center}
   \caption{Different fusion strategies. Note that, in implementation, we consider the features in both encoder and decoder.}
\label{fig:fusion}
\end{figure}

For obtained features ${\bf F}_S$ and ${\bf F}_I$, we develop an effective fusing strategy to adaptively absorb complementary information from the multi-modal contextual representations. For example, depth features encode the scene geometry structure, {\it e.g.,} the distance from the camera to partial spatial locations. It contributes to inferring the depth of unobserved locations directly. In addition, RGB features contain semantic information and provide prior appearance knowledge of unobserved pixels. Instead of concatenating or summing them together directly with or without attention mechanism, we exploit the proposed SGFM with a symmetric structure, as shown in Figure~\ref{fig:framework}. More specifically, 
at the beginning of the decoder, we employ the convolutional operation followed by a Sigmoid function on ${\bf F}_S^L$ to generate the adaptive gating weight ${\bf G}_S^L$. By applying the adaptive attention mechanism, the network can absorb meaningful information from the RGB branch and filter out the unrelated. 
Then we we feed the initial fused feature $[{\bf F}_S^L || {\bf G}_S^L\ast {\bf F}_I^L]$
into the Residual Block ({\it abbr.} ResBlock) ~\cite{he2016deep} to obtain the final fused features ${\bf Q}_{SI}^L$, which is then fed into a deconvolutional layer to generate ${\bf Q}_{SI}^{L,\uparrow}$. Therefore, the depth features can be improved by the complementary information automatically. At the other levels, there is a slight difference in learning the adaptive weights. In specific, at level $l\in\{L-1,L-2,...,0\}$, we learn the gating weights ${\bf G}_S^l$ using ${\bf Q}_{SI}^{l+1,\uparrow}$, rather than ${\bf F}_S^l$. Moreover, we feed the concatenated feature $[{\bf Q}_{SI}^{l+1,\uparrow}||[{\bf F}_S^l || {\bf G}_S^l\ast {\bf F}_I^l]]$ into the ResBlock at $l\in\{L-1,L-2,...,1\}$  or one convolutional layer at $l=0$ to get the fused feature. Due to the symmetry of the structure,  a similar procedure is employed in the image branch. To illustrate the difference between the proposed fusion strategy and the existing ones, {\it e.g.,} direct fusion and direct attention fusion, we provide the visual and quantitative comparisons among them in Figure~\ref{fig:fusion} and the ablation study, respectively.

\subsection{Branch Integration}
\label{sec:branch_fusion}
By applying the proposed symmetric gated fusion modules, we obtain two sets of features, one from the image branch and the other from the depth branch. Here, we consider two methods, {\it i.e.,} end-integration and feature-integration, to integrate them together and then obtain the final prediction result.
\subsubsection{End-integration} For each branch, we can predict a dense depth map, {\it i.e.,} $\hat{{\bf Y}}_{S},\hat{{\bf Y}}_{I}\in \mathbb{R}^{H\times W}$. Since the two branches focus on different information, the reliability of the two predictions varies across the whole image plane. To integrate them adaptively, following~\cite{qiu2018deeplidar,van2019sparse}, we further predict two confidence maps ${\bf C}_{S},{\bf C}_{I}\in \mathbb{R}^{H\times W}$, which indicate the reliability of the predictions. Therefore, the final dense depth map can be obtained as follows:
\begin{equation}
\begin{aligned}
\hat{{\bf Y}} = \frac{exp({\bf C}_{S})*\hat{{\bf Y}}_{S}+exp({\bf C}_{I})*\hat{{\bf Y}}_{I}}{exp({\bf C}_{S})+exp({\bf C}_{I})},
\end{aligned}
\label{eq:final}
\end{equation}
where $*$ represents the element-wise multiplication.
\subsubsection{Feature-integration} 
\begin{figure}[t]
\begin{center}
\includegraphics[width=0.8\linewidth,height=1.3in]{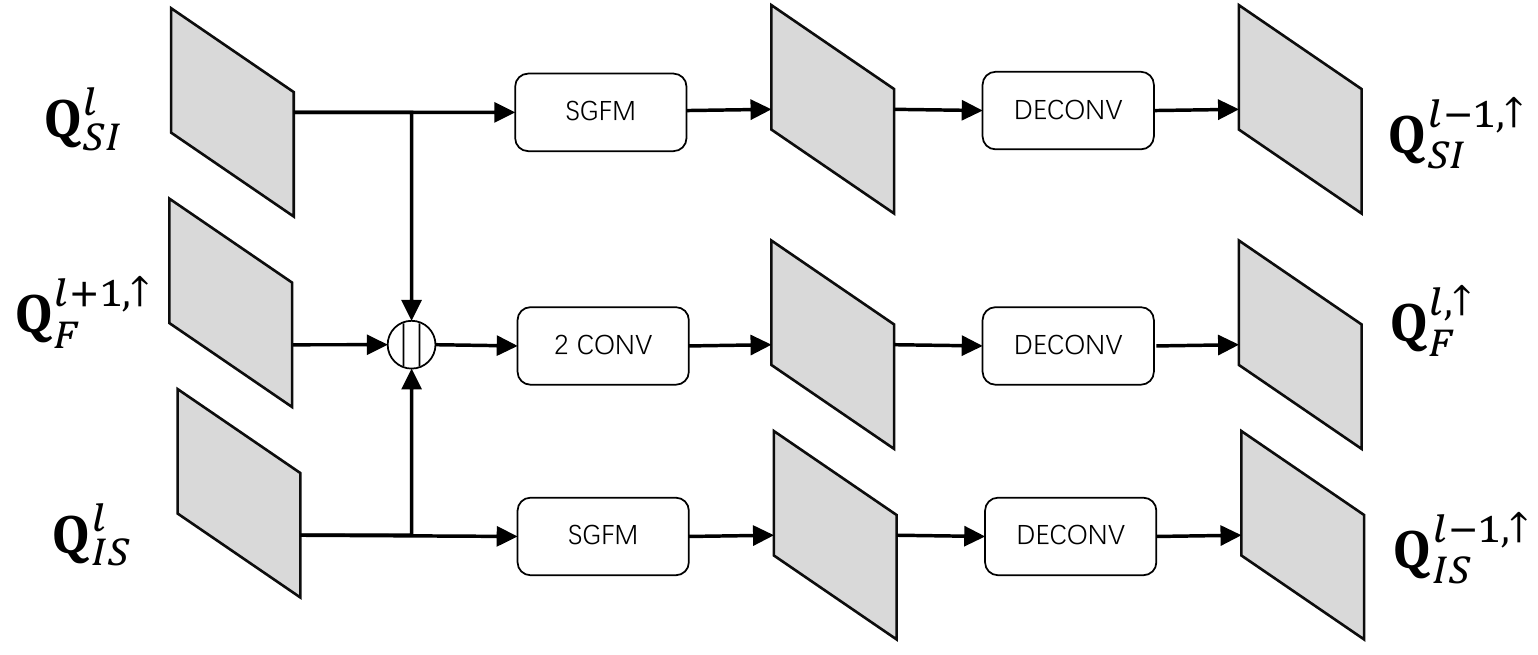}
\end{center}
   \caption{Feature-integration. Note that, we ignore some inputs of the SGFM for simplicity. }
\label{fig:integration}
\end{figure}

Apart from the integration in the end, we can also combine the features extracted by the two branches. In specific, as shown in Figure~\ref{fig:integration}, we fuse the intermediate features ${\bf Q}_{SI}$ and ${\bf Q}_{IS}$ through several convolutional operations to obtain ${\bf Q}_{F}$  progressively, and lastly obtain the final prediction $\hat{\bf Y}$ by applying one convolutional operation on the final integrated features. 

\subsection{Loss Function}
The network is mainly driven by a masked mean squared error (MSE) loss between the ground truth semi-dense depth map ${\bf Y}$ and the prediction $\hat{\bf Y}$, which is defined as:
\begin{equation}
\begin{aligned}
\mathcal{L}_{mse}({\bf Y}, \hat{\bf Y}) = \frac{1}{N_p}\sum_{i,j}\mathbb{I}({\bf Y}^{i,j}>0)({\bf Y}^{i,j}-\hat{\bf Y}^{i,j})^2,
\end{aligned}
\end{equation}
where $\mathbb{I}(\cdot)$ denotes the indication function, and $N_p$ represents the number of labeled pixels. In addition, similar to previous works~\cite{godard2017unsupervised}, we also apply an edge-aware smoothness loss to encourage depths to preserve spatial continuity:
\begin{equation}
\begin{aligned}
\mathcal{L}_{sm}(\hat{\bf Y};{\bf X}_I) = \frac{1}{N_s}||\nabla \hat{\bf Y}||_1e^{-||\nabla {\bf X}_I||_1},
\end{aligned}
\end{equation}
where $N_s$ denotes the number of pixels in the whole image space, and $\nabla$ represents first derivative along spatial directions.  Finally, the full objective is:
\begin{equation}
\begin{aligned}
\mathcal{L}(\hat{\bf Y},\hat{\bf Y}_{S},\hat{\bf Y}_{I},{\bf Y};{\bf X}_I) = &\mathcal{L}_{mse}(\hat{\bf Y},{\bf Y})+\\
&\gamma_1\mathcal{L}_{mse}(\hat{\bf Y}_{S},{\bf Y})+\\
&\gamma_1\mathcal{L}_{mse}(\hat{\bf Y}_{I},{\bf Y})+\\
&\gamma_2\mathcal{L}_{sm}(\hat{\bf Y};{\bf X}_I),
\end{aligned}
\end{equation}
where $\gamma_1$ and $\gamma_2$ are the trade-off factors, and are set to $0.5$ and $0.01$ in our experiments, respectively.

\section{Experiments}
In this section, we first introduce the datasets~\cite{Geiger2012CVPR,silberman2012indoor} used in our experiments, and
the implementation details. Then we evaluate our method by making comparisons against state-of-the-art methods. Finally, we conduct several ablations to analyze our framework.
\subsection{Benchmark Datasets}

{\bf KITTI Depth Completion Benchmark}~\cite{Geiger2012CVPR}. It is currently the main benchmark for depth completion. The dataset consists of over $90,000$ frames with the ground truth semi-dense depth map for training and validation, and $1,000$ frames without the ground-truth for test. We train depth completion models on the training set, and then evaluate the performance on the official selected validation and test sets. During training, we crop all training data (images and depth maps, $375\times 1242$) to the size of validation and test data, {\it i.e.,} $352\times 1216$. For evaluation, we adopt the official error metrics: root mean squared error (RMSE in $mm$, main metric for ranking), mean absolute error (MAE in $mm$),  root mean squared error of the inverse depth (iRMSE in $1/km$), and mean absolute error of the inverse depth (iMAE in $1/km$).

{\bf NYU-v2}~\cite{silberman2012indoor}. This dataset consists of RGB and depth images collected from $464$ different indoor scenes. According to the official data split strategy, $249$ scenes are used for training, and $654$ labeled images are selected for evaluating the final performance~\cite{eigen2014depth,laina2016deeper}. In our experiments, we sample around $48k$ images with annotations from the training set for training. Adopting similar experimental setting as~\cite{mal2018sparse,cheng2018depth}, we firstly down-sample all images to half and center-crop them to $304\times 228$, and then sample $500$ sparse LiDAR points from the provided dense depth map randomly as the sparse depth data. We exploit root mean square error (RMSE in $meter$), mean absolute relative error (REL in $meter$), and the percentage of relative errors inside a certain threshold ($\delta_{t}$, $t\in \{1.25,1.25^2,1.25^3\}$) as evaluation metrics.
\subsection{Implementation Details}

{\bf Graph Construction.} For KITTI dataset, we build the graphs at three scales with $10000$, $5000$, and $2500$ observed pixels randomly sampled from the downsampled sparse depth maps, respectively, and we calculate $6$ nearest neighbours for each node. For NYU-v2, we randomly sample $250$, $125$, and $60$ points, respectively. Note that, we can create the graphs using either the 3D coordinates ({\it e.g.,} camera coordinates) or the 2D coordinates ({\it e.g.,} pixel coordinates). Here, we use the 3D coordinates, and we will study the differences in ablation studies.

{\bf Architecture Details.} At each level of the encoder, we employ two CGPMs, and in the decoder, two ResBlocks are utilized in the symmetric gated fusion module at each scale. The feature channels in the modules are set to $64$. Our final results are obtained using the feature-integration, and in this case, we use two convolutional layers, each with $64$ output channels at each scale.

{\bf Training Details.} We implement our depth completion framework in {\it PyTorch}. In specific, we optimize our network with the momentum of $\beta_1 = 0.9$, $\beta_2 = 0.999$, and the initial learning rate of $\alpha = 0.0005$ using the ADAM solver~\cite{kingma2014adam}. The model is trained for around $40$ epochs with a batch size of 8, and
the learning rate is delayed by $0.5$ every $10$ epochs during training.

\subsection{Comparison against the State-of-the-art}
{\bf KITTI Dataset.} In Table~\ref{tb:kitti_test}, we report the number of parameters as well as the performance of our approach and previous peer-reviewed works on KITTI depth completion benchmark. Note that, some of the existing approaches employ additional data during training. For example, DeepLiDAR~\cite{qiu2018deeplidar} renders $50K$ training samples using an open urban driving simulator to train the surface normal prediction network, and Certainty~\cite{van2019sparse} utilizes a pre-trained sematic segmentation model on Cityscapes~\cite{Cordts2016Cityscapes} as network initialization, which can provide high-level semantic information for depth completion. In contrast to these approaches, we train our network from scratch without any additional data. Nevertheless, our approach obtains a convincible improvement over most of the previous methods. 
In comparison to the latest works, {\it i.e.,} CSPN++~\cite{cheng2020cspn++} and NLSPN~\cite{park2020non}, our model achieves very close performance, but our model has fewer parameters. Specifically, the RMSE errors of NLSPN and CSPN++ are $3mm$ and $1mm$ less than ours, respectively, but the number of their parameters is around four times larger than ours.

\begin{table}\scriptsize
\centering
\caption{Quantitative results on the test set of KITTI depth completion benchmark~\cite{Geiger2012CVPR}, ranked by {\bf RMSE}. The methods ranking {\color{red} first}, {\color{blue}second}, and {\color{cyan}third} are marked by the red,  blue, and cyan, respectively.
Our method performs better than most of previous methods, and yields close performance to CSPN++~\cite{cheng2020cspn++} and NLSPN~\cite{park2020non} with a much smaller model size.}
\setlength{\tabcolsep}{1.4mm}{
\begin{tabular}{l||c||c|c|c|c}
\hline
Method &  PARAM (M) & {\bf RMSE} & MAE & iRMSE & iMAE \\
\hline\hline
SparseConvs~\cite{uhrig2017sparsity} & - & 1601.33 & 481.27 & 4.94 & 1.78 \\
MorphNet~\cite{dimitrievski2018learning}& - & 1045.45 & 310.49 & 3.84 & 1.57 \\
CSPN~\cite{cheng2018depth}    & 17.41 & 1019.64 & 279.46 & 2.93 & 1.15 \\
Spade-RGBsD~\cite{jaritz2018sparse}  & $\sim$5.3 & 917.64 & 234.81 & 2.17 & 0.95 \\
HMSNet~\cite{8946876} & - & 841.78 & 253.47 & 2.73 & 1.13 \\
DDP~\cite{Yang_2019_CVPR}            & 18.8 & 832.94 & {\color{blue}203.96} & 2.10 & {\color{blue}0.85} \\
NConv-CNN-L2~\cite{eldesokey2018confidence} &  0.36 & 829.98 & 233.26 & 2.60 & 1.03 \\
Sparse2Dense~\cite{ma2018self}    & 26.10 & 814.73 & 249.95 & 2.80 & 1.21 \\
PwP~\cite{Xu_2019_ICCV}            &  28.99   & 777.05 & 235.17 & 2.42 & 1.13 \\
Certainty~\cite{van2019sparse} & 2.55 & 772.87 & 215.02 & 2.19 & 0.93 \\
DeepLiDAR~\cite{qiu2018deeplidar}  & 53.44   & 758.38 & 226.05 & 2.56 & 1.15 \\
UberATG-FuseNet~\cite{Chen_2019_ICCV} & 1.89 & 752.88 & 221.19 & 2.34 & 1.14 \\
CSPN++~\cite{cheng2020cspn++} & $\sim$26 & {\color{blue} 743.69} & 209.28 & {\color{blue}2.07} & {\color{cyan}0.90}\\
NLSPN~\cite{park2020non} & 25.84 & {\color{red}741.68} & {\color{red}199.59} & {\color{red}1.99} & {\color{red}0.84} \\
\hline
ACMNet                            & 4.9  & {\color{cyan}744.91} & {\color{cyan}206.09} & {\color{cyan}2.08} & {\color{cyan}0.90}\\
\hline
\end{tabular}
}
\label{tb:kitti_test}
\end{table}
\begin{figure*}
\begin{center}
 \includegraphics[width=1.0\linewidth,height=3.8in]{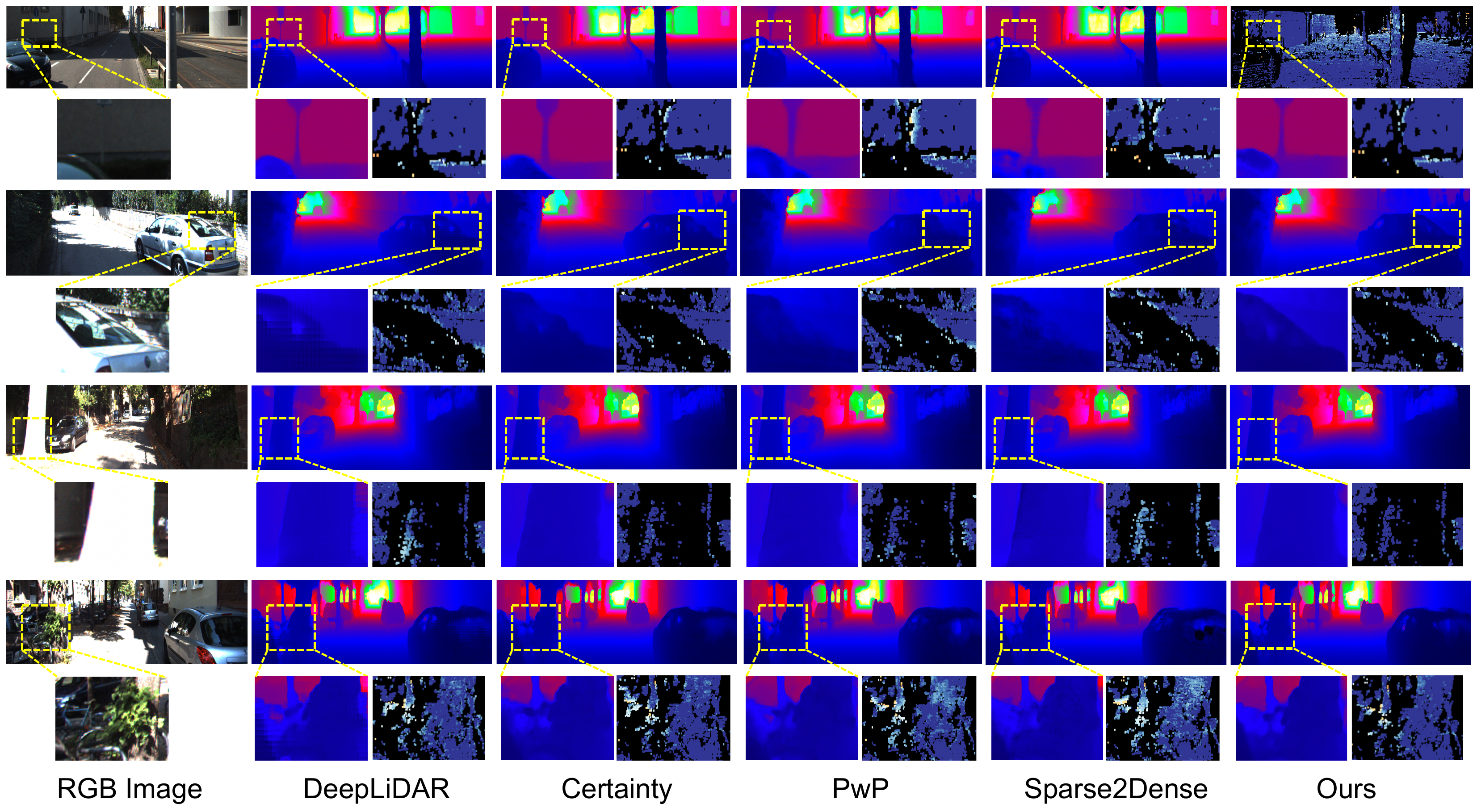}
\end{center}
   \caption{Qualitative comparison of our method against four state-of-the-art approaches on KITTI test set~\cite{Geiger2012CVPR}. Left to right: RGB image, results of DeepLiDAR~\cite{qiu2018deeplidar}, Certainty~\cite{van2019sparse}, PwP~\cite{Xu_2019_ICCV}, Sparse2dense~\cite{ma2018self}, and ACMNet, respectively. For better comparison, we show color images, dense predictions, and zoom-in views of details and error maps (darker, better). Best viewed in color.}
\label{fig:examples}
\end{figure*}
Figure~\ref{fig:examples} shows some qualitative results of ACMNet and other four state-of-the-art methods~\cite{qiu2018deeplidar,van2019sparse,Xu_2019_ICCV,ma2018self}. Benefiting from our proposed co-attention guided graph propagation and symmetric gated fusion strategy, which exploit observed pixels' information and capture the  heterogeneity of the two modalities efficiently, ACMNet is capable of yielding high-performing dense depth map, preserving more details over boundary regions ({\it e.g.,} the 2{\it nd} and 3{\it rd} examples), and  performing better on the tiny/thin objects (the 1{\it st} example).

\begin{table}\footnotesize
\begin{center}
\caption{Quantitative results on NYU-v2~\cite{silberman2012indoor} with the setting of $500$ sparse depth samples. RMSE, REL: lower better; $\delta_t$: higher better.
}
\label{tb:nyuv2}
\setlength{\tabcolsep}{2.0mm}{
\begin{tabular}{l||c|c|c|c|c}
\hline
Method &  RMSE & REL & $\delta_{1.25}$ & $\delta_{1.25^2}$ & $\delta_{1.25^3}$ \\
\hline\hline
TGV~\cite{ferstl2013image} & 0.635 & 0.123 & 81.9 & 93.0 & 96.8 \\
Bilateral~\cite{silberman2012indoor} &  0.479 & 0.084 & 92.4 & 97.6 & 98.9 \\
Zhang {\it et al.}~\cite{Zhang_2018_CVPR} & 0.228 & 0.042 & 97.1 & 99.3 & 99.7 \\
Ma {\it et al.}~\cite{mal2018sparse} & 0.204 & 0.043 & 97.8 & 99.6 & 99.9 \\
CSPN~\cite{cheng2018depth} & 0.117 & 0.016 & 99.2 &  {\bf 99.9} & {\bf 100.0} \\
DeepLiDAR~\cite{qiu2018deeplidar} & 0.115 & 0.022 & 99.3 &  {\bf 99.9} & {\bf 100.0} \\
Xu {\it et al.}~\cite{Xu_2019_ICCV} & 0.112 & 0.018 & 99.5 & {\bf 99.9} & {\bf 100.0}\\
NLSPN~\cite{park2020non} & {\bf 0.092} & {\bf 0.012} & {\bf 99.6} & {\bf 99.9} & {\bf 100.0} \\
\hline
ACMNet & 0.105 &  0.015 & 99.4 & {\bf 99.9} & {\bf 100.0}\\
\hline
\end{tabular}
}
\end{center}
\end{table}

\noindent {\bf NYU-v2 Dataset.} As shown in Table~\ref{tb:nyuv2}, most of latest works have close performance on this dataset. Our method performs better than almost all of methods except NLSPN~\cite{park2020non}, but as stated above the number of our model's parameters is far less than it.
\begin{table}\footnotesize
\centering
\caption{Quantitative results on KITTI validation set~\cite{Geiger2012CVPR} for ablation study on Graph Propagation. Noticeable improvements gained by +GP demonstrate the effectiveness of our proposed graph propagation module.}
\setlength{\tabcolsep}{3.8mm}{
\begin{tabular}{l||c|c|c|c}
\hline
Method &  RMSE & MAE & iRMSE & iMAE \\
\hline\hline
Baseline &  815.61 & 224.43 & 2.59 & 1.02 \\
+GP    &  806.87 & 220.97 & 2.42 & 0.97 \\
\hline
+GP/D  & 810.85 & 224.64 & 2.45  & 0.99 \\
+GP/W  & 809.09 & 221.44 & 2.42  & 0.97 \\
\hline\hline
+SG    & 796.79 & 219.86 & 2.39 & 0.97 \\
+GP+SG  & 789.72  & 216.65 & 2.32 & 0.96 \\
\hline
+GP/D+SG & 792.49 & 215.14 & 2.33 & 0.95 \\
+GP/W+SG & 790.75 & 217.34 & 2.39 & 0.97 \\
\hline
\end{tabular}
}
\label{tb:ab_prop}
\end{table}

\begin{table}\footnotesize
\centering
\caption{Investigation for different fusion strategies. DF: direct fusion; DAF: direct fusion with attention mechanism; SG: our proposed adaptive symmetric gated fusion strategy.}
\setlength{\tabcolsep}{4.0mm}{
\begin{tabular}{l||c|c|c|c}
\hline
Method &  RMSE & MAE & iRMSE & iMAE \\
\hline\hline
DF &  815.61 & 224.43 & 2.59 & 1.02 \\
DAF & 807.35 & 224.70 & 2.46 & 1.00 \\
SG & 796.79 & 219.86 & 2.39 & 0.97 \\
\hline
GP+DF    &  807.49 & 218.74 & 2.39 & 0.96 \\
GP+DAF & 804.69 & 221.09 & 2.44 & 0.99 \\
GP+SG  & 789.72  & 216.65 & 2.32 & 0.96 \\
\hline
\end{tabular}
}
\label{tb:ab_fusion}
\end{table}
\subsection{Ablation Study}
Here, we conduct comprehensive ablation studies on KITTI selected validation dataset to verify the effectiveness of our proposed components. In following experiments, we set the channels of intermediate layers in networks to $32$ to speed up model training. Unless otherwise specified, we exploit the end-integration in most cases.

 {\bf The effectiveness of the graph propagation.} We first demonstrate the effectiveness of the proposed co-attention guided graph propagation by comparing the performance in four cases, {\it i.e.,} (1) Baseline: no propagation used in the encoder and direct fusion in the decoder; (2) +GP: graph propagation in the encoder and direct fusion in the decoder; (3) +SG: no propagation in the encoder and symmetric gated fusion in the decoder; (4) +GP+SG: our whole model with the end-integration. As shown in Table~\ref{tb:ab_prop}, +GP and +GP+SG outperform Baseline and +SG, respectively, which 
 demonstrates that the proposed graph propagation module better captures the spatial contextual information from sparse LiDAR data. 
 
 Furthermore, we carry out four additional experiments to analyze in which stage, such as the encoder ({\it i.e.,} +GP and +GP+SG), decoder (referred as +GP/D and +GP/D+SG), or whole network (referred as +GP/W and +GP/W+SG), the graph propagation module performs better. As shown in Table~\ref{tb:ab_prop}, the comparisons (+GP {\it v.s.} +GP/D, and +GP+SG {\it v.s.} +GP/D+SG) indicate that applying the propagation module in the feature extraction stage is more effective in modeling the contextual information. Additionally, we can also observe that compared to +GP~(+GP+SG), +GP/W~(+GP/W+SG) causes some performance drop. This might be because in the decoder the structure of the observed pixels is not well-preserved after several operations in the encoder.
 
{\bf The effectiveness of the symmetric gated fusion.}
To verify that the proposed symmetric gated fusion strategy performs better than direct fusion, {\it e.g.,} concatenation with or without attention (referred as DAF and DF, respectively), we compare six models, {\it i.e.,} DF~(namely Baseline), DAF, SG, GP+DF~(namely Baseline+GP in Table~\ref{tb:ab_prop}), GP+DAF, and GP+SG. As shown in Table~\ref{tb:ab_fusion}, SG outperforms both DAF and DF, demonstrating that the proposed symmetric gated fusion strategy is capable of combining the multi-modal information more effectively. Moreover, the comparisons between GP+SG, GP+DAF, and GP+DF can further support this conclusion.

{\bf Analysis of graph construction.} Here, we investigate the impacts of three factors involved in constructing graphs. Note that, we conduct the following experiments using our final model with the end-integration. We report the results in Table~\ref{tb:ab_graph}.

Firstly, since we aim at capturing more observed multi-modal information to enhance the features of unobserved pixels by finding their spatial neighbours, it is interesting to explore the selection of the coordinate system, {\it i.e.,} pixel coordinate system or camera coordinate system. In specific, for a set of observed pixels, we can construct a graph according to their 2D coordinates  $\{(u_i,v_i)\}_{i=0}^{n-1}$ directly or 3D coordinates  $\{(x_i,y_i,z_i)\}_{i=0}^{n-1}$, which are obtained according to Eq.~\ref{eq:trans}, where $f_x,f_y,c_x,c_y$ denote the camera parameters, and $d_i$ represents the depth value. In Table~\ref{tb:ab_graph}, we compare two models (10K\_2D\_6NN {\it v.s.} 10K\_3D\_6NN), where 6-nearest neighbours algorithm is utilized to construct graphs and $10,000$ points are sampled at the first scale. We can find 10K\_3D\_6NN slightly outperforms 10K\_2D\_6NN on the RMSE metric. It is mainly because propagation in the camera (3D) coordinate system can learn the scene's geometric structure.
\begin{equation}
\begin{aligned}
&z_i = d_i\\
&x_i = \frac{z_i(u_i-c_x)}{f_x}\\
&y_i = \frac{z_i(v_i-c_y)}{f_y}\\
\end{aligned}
\label{eq:trans}
\end{equation}

Secondly, we discuss the performance of the model under different numbers of nearest neighbours. By setting $k$ ($k$ nearest neighbours) to different values, {\it i.e.,} $3,6,9$, we train three models, {\it i.e.,} 10K\_3D\_3NN ($k=3$), 10K\_3D\_6NN ($k=6$), and 10K\_3D\_9NN ($k=9$), all of which propagate features in the camera coordinate system. As shown in Table~\ref{tb:ab_graph}, in comparison to 10K\_3D\_3NN and 10K\_3D\_6NN, 10K\_3D\_9NN causes a slight decrease in the performance, it might be because increasing the number of nearest neighbours encourages the model to see unrelated contexts. 

Lastly, we study the number of sampled points. In specific, we sample $10,000$, $8,000$, and $12,000$ points at the first scale, respectively, and at the following scales, half of points are sampled from the last scale. From Table~\ref{tb:ab_graph}, we can observe that more or fewer points might degrade the performance on the RMSE metric.

In a nutshell, the selection of coordinate system, the number of nearest neighbours and sampled points might affect the performance, but in most settings, the model performs well.
\begin{table}\small
\centering
\caption{Ablation study on the coordinate system and the number of nearest neighbours and sampled points.  }
\setlength{\tabcolsep}{3.0mm}{
\begin{tabular}{c||c|c|c|c}
\hline
Graph &  RMSE & MAE & iRMSE & iMAE \\
\hline\hline
10K\_2D\_6NN &  792.56 & 216.31 & 2.34 & 0.95 \\
10K\_3D\_6NN & 789.72  & 216.65 & 2.32 & 0.96 \\
10K\_3D\_3NN & 792.13 & 216.64 & 2.35 & 0.96 \\
10K\_3D\_9NN & 795.09 & 216.57 & 2.37 & 0.96 \\
08K\_3D\_6NN & 794.59 & 216.64 & 2.36 & 0.95 \\
12K\_3D\_6NN & 793.61 & 215.81 & 2.34 & 0.95 \\
\hline
\end{tabular}
}
\label{tb:ab_graph}
\end{table}

\begin{figure*}
\begin{center}
 \includegraphics[width=0.8\linewidth,height=2.0in]{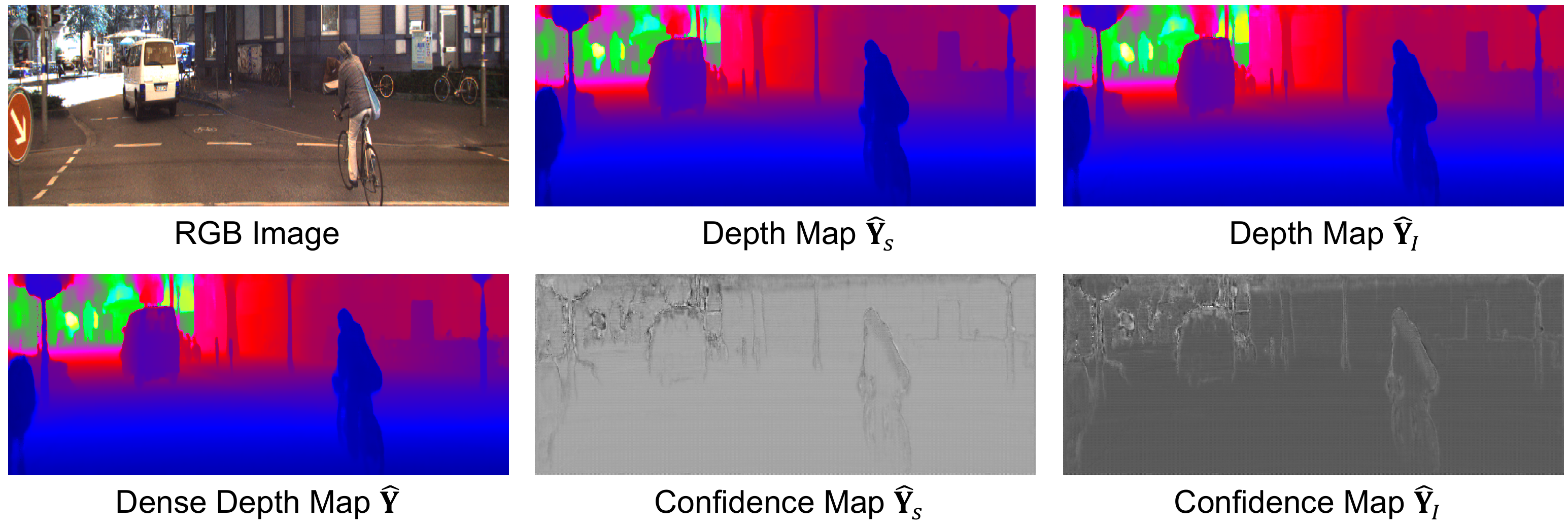}
\end{center}
   \caption{Qualitative example of the end-integration. First row: input image, prediction of EI/Depth and EI/Image, respectively; Second row: final prediction, and confidence maps corresponding to the predictions in the first row. We can find that each branch can capture different information.}
\label{fig:two_branch}
\end{figure*}
{\bf Analysis of branch integration.}
In Section~\ref{sec:branch_fusion}, we introduce two methods for the integration of the two branches. Here, we analyze their performances.  As shown in Table~\ref{tb:ab_integration}, the comparison (RMSE: 786 {\it v.s.} 789) between Feature-Integration ({\it abbr.} FI) and End-Integration ({\it abbr.} EI) shows that integration at the feature level is more powerful than the end in learning the reliability of the two branches.

In addition, we also evaluate the performance of the two branches. Taking the end-integration as an example, we report the performance of EI/Depth fusing the RGB information into the depth, and EI/Image doing the opposite. Although the two branches yield close scores on all metrics, by learning confidence maps to fuse them together, a significant improvement on all metrics is obtained. To understand the two branches deeply, we provide a qualitative example in Figure~\ref{fig:two_branch}. It can be seen that the depth branch is able to generate dense depth map with higher confidence in most locations, while the image branch performs better in capturing the boundary information. This result also further supports that the two modalities are  complementary to each other.
\begin{table}\footnotesize
\centering
\caption{Investigation for the two proposed integration methods.}
\setlength{\tabcolsep}{2.8mm}{
\begin{tabular}{c||c|c|c|c}
\hline
Method &  RMSE & MAE & iRMSE & iMAE \\
\hline\hline
End-Integration & 789.72  & 216.65 & 2.32 & 0.96 \\
Feature-Integration & 786.89 & 216.24 & 2.28 & 0.96 \\
\hline
EI/Depth &  802.66 & 219.88 & 2.40 & 0.97 \\
EI/Image & 807.34  & 223.26 & 2.47 & 1.00 \\  
\hline
\end{tabular}
}
\label{tb:ab_integration}
\end{table}

\subsection{Generalization Capabilities on Different Levels of Sparsity}
\begin{figure*}
\centering
\captionsetup[sub]{font={scriptsize}}
   \begin{subfigure}[b]{0.30\textwidth}
    \centering
     \includegraphics[width=1\linewidth,height=1.9in]{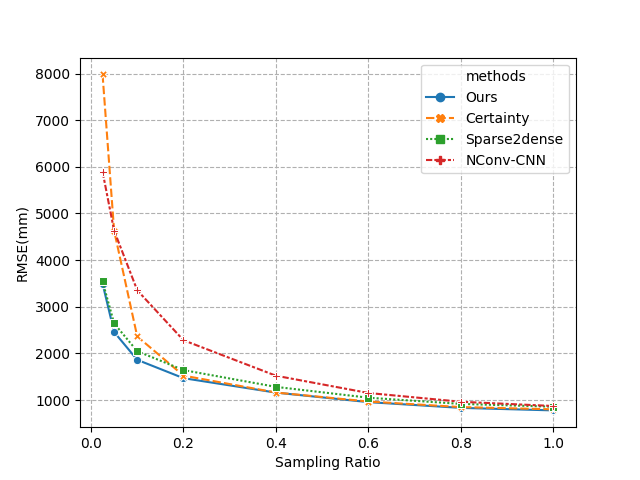}
      \subcaption*{RMSE on All Sampling Ratios}
    \end{subfigure}%
    \begin{subfigure}[b]{0.30\textwidth}
    \centering
     \includegraphics[width=1\linewidth,height=1.9in]{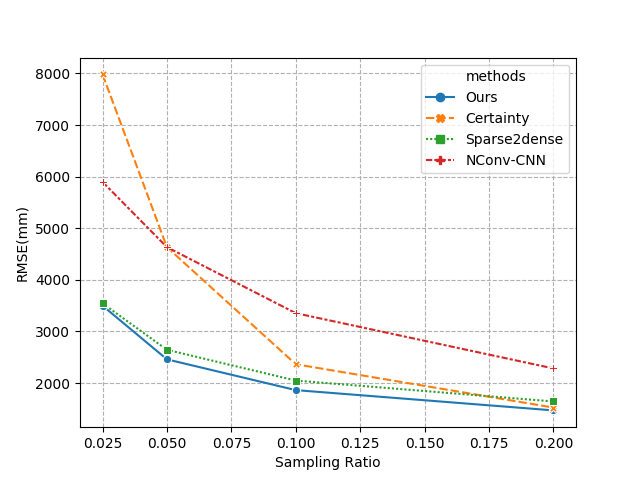}
      \subcaption*{RMSE on Sampling Ratios $0.025, 0.05, 0.1,0.2$ }
    \end{subfigure}%
    \begin{subfigure}[b]{0.30\textwidth}
    \centering
     \includegraphics[width=1\linewidth,height=1.9in]{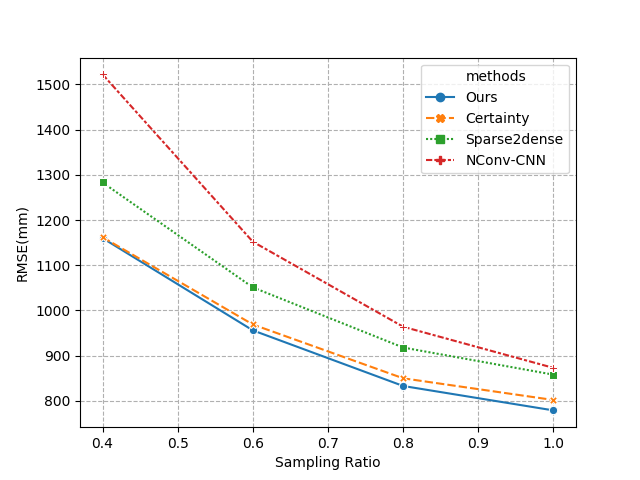}\\
      \subcaption*{RMSE on Sampling Ratios $0.4, 0.6, 0.8,1.0$ }
    \end{subfigure}

    \begin{subfigure}[b]{0.30\textwidth}
    \centering
     \includegraphics[width=1\linewidth,height=1.9in]{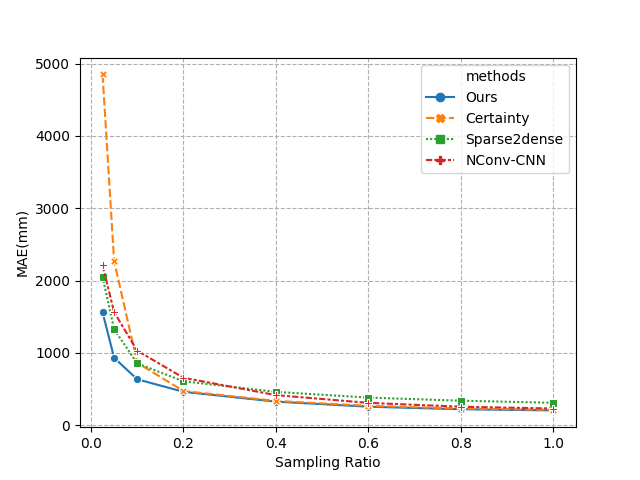}
      \subcaption*{MAE on All Sampling Ratios}
    \end{subfigure}
    \begin{subfigure}[b]{0.30\textwidth}
    \centering
      \includegraphics[width=1\linewidth,height=1.9in]{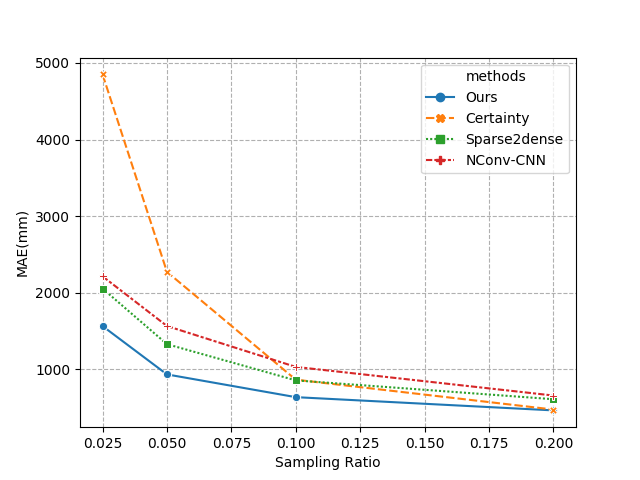}
      \subcaption*{MAE on Sampling Ratios $0.025, 0.05, 0.1,0.2$}
    \end{subfigure}
    \begin{subfigure}[b]{0.30\textwidth}
    \centering
      \includegraphics[width=1\linewidth,height=1.9in]{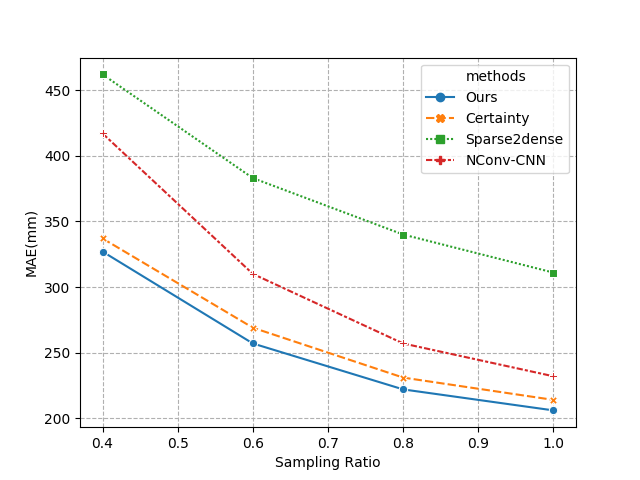}
      \subcaption*{MAE on Sampling Ratios $0.4, 0.6, 0.8,1.0$}
    \end{subfigure}
  \caption{Performances under different levels of sparsity. For better comparison, we also show the performances on lower and larger densities separately in the right two figures of each row. In comparison to Certainty~\cite{van2019sparse}, Sparse2dense~\cite{ma2018self}, and NConv-CNN~\cite{eldesokey2018confidence}, ACMNet performs better under all input densities.}
  \label{fig:sparsity} 
\end{figure*}

To show the generalization capabilities of ACMNet on different levels of sparsity, we evaluate our approach and other three state-of-the-art methods with publicly available code, {\it i.e.}, Certainty~\cite{van2019sparse}, Sparse2dense~\cite{ma2018self}, and NConv-CNN~\cite{eldesokey2018confidence}, on KITTI selected validation set under different input densities. In specific,
we first uniformly sub-sample the raw LiDAR depth by ratios of $0.8, 0.6, 0.4, 0.2, 0.1, 0.05$, and $0.025$ to generate sparse depth maps with different densities, and then test pretrained models on the generated sparse depth maps. Note that, all the models are trained on KITTI training set under the original sparsity (sampling ratio of $1.0$) but not fine-tuned on the new sparse depth maps. Figure~\ref{fig:sparsity} shows that our approach performs better under all input densities in terms of both RMSE and MAE metrics, which demonstrates the impressive generalization capabilities of our approach under different levels of sparsity.

\section{Conclusion}
In this paper, we have developed an Adaptive Context-Aware Multi-Modal Network (ACMNet) to recover a dense depth map from sparse LiDAR data and dense RGB data. The critical issue in depth completion is 
how to exploit the observed spatial contexts from multi-modal data efficiently. To this end, we apply the co-attention guided graph propagation within multiple graphs constructed from observed pixels, which adaptively extracts multi-scale and multi-modal features and contributes to the feature enhancement for unobserved pixels. Furthermore, to fuse the multi-modal features in an effective way, we propose the symmetric gated fusion strategy, which has the capability of learning the heterogeneity of the two modalities. Finally, we implement our ACMNet, where a stack of CGPMs are employed in the encoder and SGFMs are used in the decoder.
 Benefiting from the two new modules, ACMNet is capable of generating high-quality dense depth maps.
Our extensive experiments have demonstrated the effectiveness of the network as well as the network components.

{\small
\bibliographystyle{IEEEtran}
\bibliography{egbib}
}

\end{document}